\documentclass[letterpaper, 10 pt, conference]{ieeeconf}  %

\IEEEoverridecommandlockouts                              %

\usepackage{xcolor}
\usepackage{url}
\usepackage{amsmath}
\usepackage{amssymb}  
\usepackage{color}  
\usepackage{graphicx}  
\usepackage{pifont}

\usepackage{booktabs}  
\usepackage{multirow}  

\usepackage[pagebackref=true,breaklinks=true,letterpaper=true,colorlinks,bookmarks=false]{hyperref}

\definecolor{lightgray}{RGB}{220,220,220}
\definecolor{darkblue}{RGB}{0,0,127}
\definecolor{darkgreen}{RGB}{0,127,0}
\definecolor{darkred}{RGB}{200,0,0}

\usepackage{array}
\newcolumntype{L}[1]{>{\raggedright\let\newline\\\arraybackslash\hspace{0pt}}m{#1}}
\newcolumntype{C}[1]{>{\centering\let\newline\\\arraybackslash\hspace{0pt}}m{#1}}
\newcolumntype{R}[1]{>{\raggedleft\let\newline\\\arraybackslash\hspace{0pt}}m{#1}}

\def\ie{\emph{i.e.,}~}
\def\eg{\emph{e.g.,}~}

\def\viz{\emph{viz.,}~}

\def\etal{\emph{et~al.}} 

\title{\LARGE \bf
Indirect Object-to-Robot Pose Estimation \\ from an External Monocular RGB Camera
}

\author{Jonathan Tremblay \hspace{2em} Stephen Tyree \hspace{2em} Terry Mosier \hspace{2em} Stan Birchfield \\
NVIDIA
}

\begin{document}

\maketitle
\thispagestyle{empty}
\pagestyle{empty}

\begin{abstract}

We present a robotic grasping system that uses a single external monocular RGB camera as input.
The object-to-robot pose is computed indirectly by combining the output of two neural networks:  one that estimates the object-to-camera pose, and another that estimates the robot-to-camera pose.
Both networks are trained entirely on synthetic data, relying on domain randomization to bridge the sim-to-real gap.
Because the latter network performs online camera calibration, the camera can be moved freely during execution without affecting the quality of the grasp.
Experimental results analyze the effect of camera placement, image resolution, and pose refinement in the context of grasping several household objects.
We also present results on a new set of 28 textured household toy grocery objects, which have been selected to be accessible to other researchers. To aid reproducibility of the research, we offer 3D scanned textured models, along with pre-trained weights for pose estimation.\footnote{Project page  \href{https://research.nvidia.com/publication/2020-07_Indirect-Object-Pose}{link}.  Video: \url{https://youtu.be/E0J91llX-ys}}
\end{abstract}

\section{INTRODUCTION}

Determining the pose of an object \emph{with respect to the camera} has received much attention in the computer vision and robotics communities~\cite{xiang2018rss:posecnn,tremblay2018corl:dope,peng2019pvnet,zakharov2019dpod,hu2019segmentation,wang2019cvpr:densefus}.
However, this is only half the battle.
To be useful in the context of a robotic manipulation system, the object's pose must be ascertained \emph{with respect to the robot}.

The standard setup involves either 1) an in-hand camera, or 2) an external camera.
The former approach has the advantage that, once the camera's pose has been calibrated with respect to the robot's end effector, it never needs to be recalibrated (assuming that it remains rigidly attached).  
Moreover, a camera-in-hand is able to see the object at arbitrarily close distances and multiple viewpoints, thus providing a potentially large signal-to-noise ratio to determine the object's location before grasping.
On the other hand, a camera-in-hand has difficulty seeing the surrounding context; if, for example, the object is disturbed by an external force that moves it out of the camera's field of view, the robot must blindly recover---a potentially dangerous maneuver when the environment is dynamic and unstructured.
As a result, camera-in-hand systems are often augmented with an external camera for sensing the robot's workspace from a global view.

This reasoning leads naturally to the latter approach of primarily using an external camera for pose estimation.
However, external cameras typically require an off-line calibration process that is notoriously tedious, sensitive, and error-prone.
This process computes the camera's pose with respect to the robot in one or more images using fiducial markers such as ARTags~\cite{fiala2005cvpr:artag} or AprilTags~\cite{olson2011icra:apriltags}.
Such markers must be placed carefully to avoid occlusion, and their pose with respect to the robot must be known.
Moreover, once the camera has been calibrated in this manner, the camera's pose cannot be disturbed, or the entire delicate procedure must be repeated.

\begin{figure}
    \centering
    \begin{tabular}{c}
        \hspace{-0.5em}\includegraphics[width=0.95\linewidth]{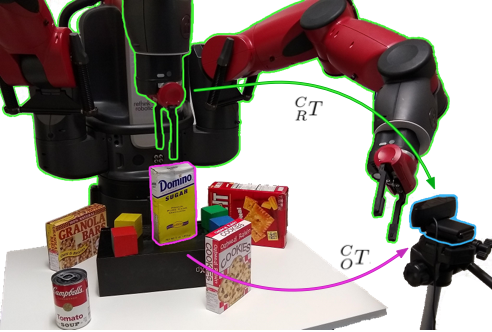}
    \end{tabular}
    \caption{A Baxter robot is controlled solely by an external monocular RGB camera mounted on a tripod.  Images from the camera are used to \emph{indirectly} compute the pose of the object (\viz the sugar box) with respect to the robot, for the purpose of grasping.  On-line calibration allows the camera to be moved while the robot is running.}
    \label{fig:setup}
\end{figure}

One of the goals of our lab is to develop a system that can reliably and robustly perform pick-and-place of known objects in its workspace using readily-available sensing technology.
In this paper we report on recent progress toward that goal, in which we equip a robot with a single monocular RGB camera mounted on an external tripod.
As shown in Fig.~\ref{fig:setup}, an object's pose with respect to the robot is computed indirectly by combining the output of two neural networks:
one that estimates the object's pose with respect to the camera~\cite{tremblay2018corl:dope}, and another that estimates the camera's pose with respect to the robot~\cite{lee2020icra:dream}.
The latter network enables online calibration, so the camera can be moved freely without affecting the quality of the results.
These networks were both developed in our lab and were trained using only synthetic data, relying on domain randomization~\cite{tobin2017iros:dr} to bridge the sim-to-real gap.

This paper contains the following contributions:
\begin{itemize}
    \item We present a robotic system capable of estimating the pose of known objects \emph{in the robot's coordinate system} using an external monocular RGB camera with online calibration.
    \item The accuracy of the system is analyzed, along with the effects of image resolution, camera position, pose refinement, and other system parameters.  We show that accuracy of $2$~cm is achievable under a variety of conditions.
    \item A robotic grasping system is demonstrated, showing the ability to grasp objects even when the camera is moved.  These objects include a set of YCB objects, along with a novel set of $28$ readily acquired toy grocery objects which we introduce.  To aid reproducibility, we provide 3D scanned textured models and pre-trained pose-estimation weights for these new objects.
\end{itemize}

\section{PREVIOUS WORK}

One useful way to categorize robotic grasping systems is by the type of sensor input.
Many modern robotic manipulation systems rely on commodity RGBD sensors (\eg Kinect, RealSense, or PrimeSense).
One approach is to use the point cloud to refine the pose of an object once it has been detected, as in PoseCNN~\cite{xiang2018rss:posecnn}. 
A more principled approach is to learn to jointly process depth and RGB to estimate the object's pose, as in DenseFusion~\cite{wang2019cvpr:densefus}. 
The same researchers~\cite{wang2020icra:6pack} show that it is possible to learn to extract 3D keypoint representations from RGBD data, which can then be used for robotic manipulations, \eg pouring.
Stev\v{s}i\'{c} \etal~\cite{stevsic2020ral:ltass} use RGBD pose estimation for an assembly task, focusing on detecting poses of known geometries in arbitrary and complex contexts.
The promising works of Florence \etal~\cite{florencemanuelli2018dense} and Manuelli \etal~\cite{manuelli2019kpam} show it is possible to learn feature representations (sparse or dense, respectively) from RGBD inputs which are consistent across instances in the same category (\eg mugs) and useful for grasping and placing objects in desired poses.
The former approach, which operates on keypoints, shares similarities with that of our earlier work in Tremblay \etal~\cite{tremblay2018icra:cube}.

Another approach is to use the point cloud from the depth camera to predict grasping positions directly, without RGB information.
Mahler~\etal~\cite{mahler2017rss:dexnet2} learn to evaluate the quality of candidate top-down grasps sampled from point clouds.
Mousavian~\etal~\cite{Mousavian_2019_ICCV} learn to sample and refine a diverse set of $6$-DoF (degrees-of-freedom) grasps extracted from segmented point clouds.
Murali~\etal~\cite{murali2020icra:cluttergrasping} extended this work to cluttered scenarios by learning to discriminate between successful grasps and grasps which collide with the environment. 

Finally, a few systems rely on a single external RGB camera. 
Tremblay \etal~\cite{tremblay2018corl:dope} present a method to retrieve the $6$-DoF pose of known household objects from a single RGB camera which has been previously calibrated to the robot.
The method leverages a network to predict 2D image coordinates for corners of a bounding cuboid of known dimensions, from which the pose is computed using P$n$P.
Pauwels and Kragic~\cite{pauwels2016icra:onlinecalib} present an online calibration procedure that tracks objects to compute the camera-to-robot pose from multiple images.
Other work leverages an RGB camera to learn a policy in image space to control a robotic arm for grasping. 
For example, Tobin \etal~\cite{tobin2017iros:dr}, James \etal~\cite{james2017corl:transferring}, and Rusu \etal~\cite{rusu2017corl:prog} use simulation to learn neural network policies capable of directing a robotic arm to grasp colored cubes (or other simple geometric shapes) using images as input.
These works do not separately compute the pose of the object or robot, and have demonstrated only limited tolerance to variation in camera pose or workspace configuration (\eg dimensions of table, height of object) relative to training.
Similar approaches have also been explored in the context of reinforcement learning~\cite{lee2020icra:guapo,smith2019avid,iqbal2020icra:directional}.
To date, no work has shown the ability to grasp objects using a single RGB camera with online extrinsic calibration.

\section{METHOD}

In this section we describe in detail the different components of our proposed system.

\subsection{Estimating object-to-camera pose}

We use a deep neural network to estimate the $6$-DoF pose (\ie 3D rotation and translation) of known objects with respect to the camera.
This network, based upon our earlier work called DOPE~\cite{tremblay2018corl:dope}, consists of a multiple stage convolutional network that transforms the input RGB image into a set of belief maps, one per keypoint.
Typically, $n=9$ keypoints are used to represent the vertices of a bounding cuboid, along with the centroid.
In addition to the belief maps, the network outputs $n\!-\!1$ affinity maps, one for each non-centroid keypoint.
Each map is a 2D field of unit vectors pointing toward the nearest object centroid.
The maps are used by a postprocessing step to individuate objects, allowing the system to handle multiple instances of each object type.
Pose is determined by applying a P$n$P algorithm~\cite{lepetit2009ijcv:epnp} to the keypoints detected as peaks in the belief maps.

The input to the network consists of $533\!\times\!400$ images processed by a VGG-based~\cite{simonyan2015iclr:vgg} feature extractor, resulting in a $50\!\times 50\!\times\!512$ block of features.
These features are processed by a series of $6$ stages---each with $7$ convolutional layers---which output and refine the belief maps described above.
Please refer to \cite{tremblay2018corl:dope} for further details.
For training, we use $120$K synthetic images of each object generated by our publicly-available NDDS tool~\cite{to2018ndds}.
For the YCB objects, the dataset was split between domain-randomized~\cite{tobin2017iros:dr} images and photorealistic images from FAT~\cite{tremblay2018arx:fat}, as described in Tremblay \etal~\cite{tremblay2018corl:dope}; for the new objects, all data was domain-randomized, without photorealistic data.
Training consisted of $60$ epochs using the Adam optimizer~\cite{kingma2015iclr:adam} and a learning rate of $10^{-4}$. %

\subsection{Estimating camera-to-robot pose}

The $6$-DoF pose of the robot with respect to the camera is also estimated using a deep neural network.
This network, based upon our recent work called DREAM~\cite{lee2020icra:dream}, consists of an encoder-decoder that transforms the input RGB image into a set of belief maps, one per keypoint.
Because only one robot is in the scene, affinity fields are not needed.
In this paper we focus on the Rethink Robotics Baxter bimanual robot with which our lab is equipped.
While this robot was mentioned in Lee \etal~\cite{lee2020icra:dream}, no quantitative experiments were performed.
This paper extends that work by analyzing the performance quantitatively in the context of object grasping, along with various system improvements that we found necessary to achieve good results in practice.

\begin{figure}
    \centering
    \begin{tabular}{c}
        \hspace{-0.5em}\includegraphics[width=0.75\linewidth]{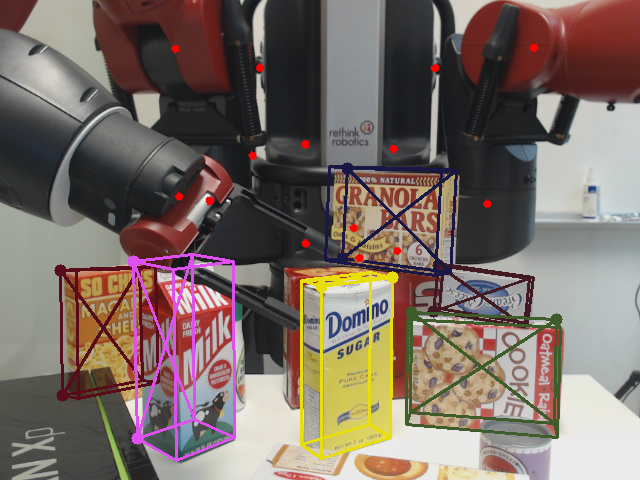}
    \end{tabular}
    \caption{The scene from the external camera's point of view, with overlaid detections of the objects (colored cuboids) and robot keypoints (red dots) found by the two networks.  Note that the networks are able to leverage the surrounding context to detect keypoints despite severe occlusion.}
    \label{fig:baxter_detections}
\end{figure}

One of those improvements is to augment the keypoints located at the $14$ joints of the robot with $10$ additional keypoints defined manually on the robot's torso (see Fig.~\ref{fig:baxter_detections}).
These latter keypoints are important for achieving pose stability when the arms are mostly out of the camera's field of view, which occurs when the camera is viewing the scene from a close range (less than about $1.4$~m).
The input to the network is a $400\!\times\!400$ image, downsampled and center cropped from the original $640\!\times\!480$.
The layers of the network are as follows: the encoder follows the same layer structure as VGG, whereas the decoder 
is constructed from $4$ upsample layers followed by $2$ convolution layers resulting in the keypoint belief maps.
The network is trained using synthetic data generated using an internal version of NDDS~\cite{to2018ndds} that allows the robot joints to be moved at random while simultaneously randomizing the camera pose.
As with DOPE, domain randomization~\cite{tobin2017iros:dr} is used in training to bridge the sim-to-real gap.
In this work we also propose a new version of the architecture that uses two stages to further refine the estimates. 

\subsection{Estimating object-to-robot pose}

For grasping an object, it is imperative to calculate the pose of the object with respect to the robot's coordinate frame.
This is achieved simply as:
\begin{equation}
{^R_O}T = \left({^C_R}T\right)^{-1} \, {^C_O}T,
\label{eq:main}
\end{equation}
where ${^R_O}T$ is the pose of the object in the robot frame, ${^C_R}T$ is the pose of the robot in the camera frame (calculated by the second network), and ${^C_O}T$ is the pose of the object in the camera frame (calculated by the first network).

Note that in contrast to end-to-end learning~\cite{tobin2017iros:dr,james2017corl:transferring,jang2017corl:semgrasp,xu2018icra:ntp,rusu2017corl:prog,levine2016jmlr:ete,zhu2018rss:rlildvs}, the modular approach presented here has the advantage that the system can be repurposed without having to retrain all the networks.  
For example, to apply our system to a new robot, only one of the networks needs to be retrained.
Similarly, to apply the system to new objects, networks only need to be trained for those objects, independent of the robot and of other objects.
Moreover, this modular approach facilitates testing and refinement of the individual components to ensure accuracy and reliability.

\subsection{Pose refinement}

Estimating the $6$-DoF pose of an object from a monocular RGB camera is challenging, because---being an inverse problem---small errors in the input can lead to large errors in the output.
Any network that directly outputs keypoint locations or poses will create errors in the estimate due to limited training data and network capacity.
One way to address this issue is to apply a pose refinement step that adjusts the pose parameters by iteratively matching the input image with a synthetic projection of the model according to the current pose.
For this step we rely upon the matching technique of Pauwels \etal~\cite{pauwels2016csvt:posetrack,pauwels2015iros:simtrack}, which we modified by replacing the detection step with the first network above; thus only the tracking step is used for pose refinement.
This additional processing helps to insulate measurements from slight errors caused by the networks.
\section{TOY OBJECT SET}

Reproducibility and sharing solutions to accelerate the pace of research are fundamental challenges for the robotics research community.
Given the progress in computer vision over the past decade, we aim to replace fiducial markers with unmodified, everyday objects for robotic manipulation research, thereby allowing more realistic scenarios.
The most widely used set of objects in manipulation research today are the YCB objects~\cite{calli2015ram:ycb}.
However, these objects are no longer easily acquired\footnote{Even for researchers in the USA, it is difficult to purchase matching items due to lack of continued availability or significant changes in texture; in other countries the problem is even more acute.}
and only a subset meets the size and weight constraints for manipulation with typical robotic arms.

To this end, we propose a new set of objects for robotic manipulation research.
Our design objectives dictate that the set of objects be:  1) somewhat realistic, 2) of the proper size and shape for grasping by typical robotic end effectors, and 3) accessible to researchers throughout the world.  
One important emerging area in robotic manipulation is that of household robots for automating daily chores in, say, a kitchen.  
Such applications are crucial for facilitating aging-in-place, assisted living, and healthcare-related challenges.
While we initially intended to scan real items from a local grocery store, we quickly realized that this approach comes with several fundamental limitations:  1) such objects are not widely available to researchers around the world, 2) the reflective properties of many of these objects (\eg metallic cans or reflective labels) make for difficult scanning, and 3) the texture of real-world objects frequently change due to seasonal marketing campaigns.

\begin{figure}
    \centering
    \begin{tabular}{c}
        \hspace{-0.5em}\includegraphics[width=0.95\linewidth]{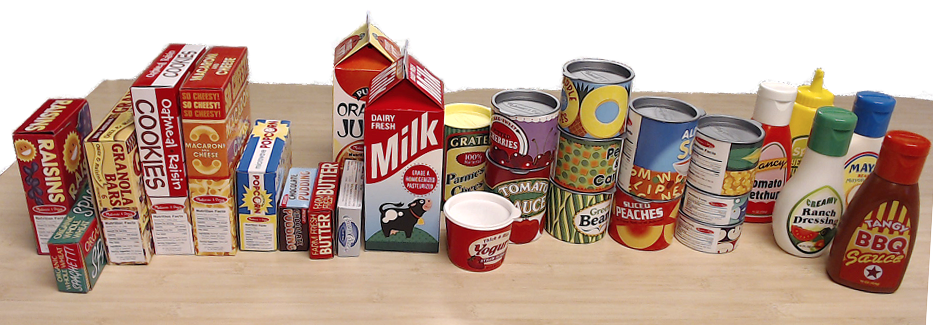}
    \end{tabular}
    \caption{Set of $28$ toy grocery objects used in the experiments.}
    \label{fig:objects}
\end{figure}

As a result, we settled on a set of $28$ toy grocery objects that are widely available online, see Fig.~\ref{fig:objects}.
Because the containers are hollow, there are no issues with perishability or shipping, and therefore they can be easily purchased online by anyone in the world.
Total cost for all objects is less than $50$ USD.  
Moreover, these objects are of the proper size and shape for typical robotic grippers.
Using an EinScan-SE scanner, we generated 3D textured models of these objects, from which we created synthetic training images for pose estimation neural networks.
These 3D models and the trained pose estimation weights for all $28$ objects are made available, both as a baseline system for comparison as well as for practical use in robotics labs.

\section{EXPERIMENTAL RESULTS}

In this section we seek to answer the following questions: 
What absolute accuracy can the system achieve in estimating the object-to-robot pose?
How is accuracy affected by image resolution, robot keypoints, and pose refinement strategies?
When the camera is moved around the scene, how accurately can the system servo above an object?
And, finally, how does object-to-robot pose accuracy translate into grasp success?

\subsection{Experimental setup}

Our experiments use a Baxter robot which has two $7$-DoF (degrees of freedom) arms; 
we use only the right arm.
The parallel jaw gripper's range of motion is $4$~cm, so pose estimates with error greater than $2$~cm generally result in a failed grasp.\footnote{
More specifically, for an upright cuboid, this threshold is only applicable along the line joining the two fingers, as the system is more forgiving in the orthogonal direction.  
For an upright cylinder, the threshold is applicable in both directions.
In all cases, the system is more forgiving along the approach direction (vertical in our experiments).}
The system uses a single Logitech C920 webcam, and all processing is done on a local PC with an NVIDIA Titan Xp GPU.
Given the reach of the arm and field of view of the camera, the robot's workspace, \ie the volume in which graspable objects can be placed, is about $20\!\times\!20\!\times\!15$~cm in front of the robot. 
The camera can be placed anywhere in front of the robot within approximately $90^\circ$ azimuth, distance from $1$ to $2$~m, and height approximately the same as the torso.
Although full $6$-DoF pose estimation facilitates more complex grasping maneuvers, for this paper we limit the system to using top-down grasps for simplicity; the extension to other grasps is straightforward.
Even so, note that grasps can occur at arbitrary height (\ie objects are not restricted to rest on the table or any other plane).

The robot is controlled using Riemannian Motion Policies (RMPs)~\cite{ratliff2018arx:rmp,cheng2018wafr:rmpflow} to follow pre-specified waypoints 
(relative to the object) to move the robot to different locations.
Both neural networks process a single image, with no tracking over time.
As a result, for safety and stability, both object-to-camera and camera-to-robot poses are processed with an exponentially weighted filter to dampen oscillations that otherwise would occur from errors in the network outputs.
This introduces a slight lag (on the order of $1$--$2$ seconds) in system response.

\subsection{Positioning accuracy}

We assessed the accuracy of the pose estimation of the object with respect to the robot for a variety of objects, camera positions, and system configurations.
We focused our measurements on the horizontal $x$--$y$ plane only, since this is where the tolerance is most important for a top-down grasp.
For setup, we removed the robot's gripper, pointed the wrist downward, and hung a weight from a string attached to the middle of the wrist.
Gravity held the weight directly below the wrist, allowing manual measurement of the lateral distance in the horizontal plane between the center of the wrist and the center of the object. 
If the system were perfect, the dangling weight would align laterally with the center of the top face of the object.

For this first set of experiments, we explored two design choices using the sugar box from the YCB objects~\cite{calli2015ram:ycb} (yellow outline in Fig.~\ref{fig:baxter_detections}).
The object was placed in four locations around the table, in an approximately $20$~cm by $20$~cm area, depending upon the camera's field of view and the robot's reach. 
The object was also placed with an in-plane rotation of approximately $45^\circ$ to ensure that the camera could see two faces of the object.
Before manually taking a measurement, we waited several seconds to let the system stabilize, to reduce the effects of the exponential filtering.
Mean error at these locations was calculated for each configuration.

Tab.~\ref{tab:expdist} shows results of this experiment.
The first column (``network'') details two design choices of the robot-to-camera network: whether the output dimensions are $100\!\times\!100$ or $400\!\times\!400$, and whether joints ($J$) and/or torso ($T$) keypoints were used.
The dimensions of the object-to-camera network output were fixed to $50\!\times\!50$.
The second column (``ref'') indicates whether the pose refinement was used.
Error is measured at a range of camera-to-robot distances ($d_i$) from $102$ to $162$~cm, \ie $d_i=92+10i$~cm.
At these distances, the width of the sugar box in the $640\!\times\!480$ image varies from approximately $99$ to $51$~pixels.
Due to space limitations, only mean error is shown.

From the first row, we see that the baseline DREAM network from~\cite{lee2020icra:dream}, which was trained only on the robot joints, does not have sufficient information to estimate the camera-to-robot pose at distances closer than $152$~cm; and when the camera is far, a sufficient number of joints are seen, but the accuracy is poor.
The next four rows show the effects of image resolution and pose refinement on our newly trained versions, both of which help to improve accuracy.
Interestingly, the penultimate row shows that higher-resolution output with fewer keypoints ($400\!\times\!400,T$) yields similar results to 
smaller-resolution output with more keypoints ($100\!\times\!100,JT$). 
Note that with both joints and torso keypoints ($JT$), performance is more consistent across distances:  closeup, the torso keypoints are visible, whereas farther away, both joint and torso keypoints are visible, which helps to mitigate errors which otherwise would be expected from increasing distance.

\hspace{-0.9em}
\begin{table}
\caption{Robot-to-object positioning error under different system configurations and varied distance ($d_i$) from camera to robot, with YCB sugar box.}
\label{tab:expdist}
\centering
\begin{small}
\begin{tabular}{ccccccccc}
\toprule
\multicolumn{2}{c}{system} & \multicolumn{7}{c}{mean error (cm), $4$ object positions} \\
\cmidrule(r){1-2} \cmidrule(r){3-9} 
network & ref & $d_1$ & $d_2$ & $d_3$ & $d_4$ & $d_5$ & $d_6$ & $d_7$ \\
\midrule
$100,J$ & \checkmark & - & - & - & - & - & 5.5 & 3.3 \\
$100,T$ & & 0.7 & 1.8 & 2.0 & 3.0 & 4.0 & 4.8 & 8.0 \\
$100,T$ & \checkmark & 0.7 & 1.0 & 1.7 & 1.7 & 2.4 & 2.5 & 3.6 \\
$400,T$ & & 1.3 & 0.8 & 0.9 & 1.6 & 1.5 & 2.8 & 3.6 \\
$400,T$ & \checkmark & 0.7 & \textbf{0.7} & 0.8 & \textbf{0.9} & 1.4 & 2.6 & 2.0 \\
$100,JT$ & \checkmark & \textbf{0.6} & \textbf{0.7} & \textbf{0.5} & 1.3 & \textbf{1.3} & \textbf{1.1} & \textbf{1.9} \\
\bottomrule
\end{tabular}
\end{small}
\end{table}

In the next experiment, shown in Tab.~\ref{tab:expdist_ycb}, we evaluated the performance of the system for $5$ YCB objects.
Objects were placed (as before) at a favorable angle to allow the perception system to view multiple sides.
For this experiment we used the $100\!\times\!100, JT$ network and included pose refinement.
For most objects, the errors are within the threshold required for grasping at most distances.
The low accuracy of the mustard was likely caused by the lack of texture on the object, which caused the pose refinement to inaccurately estimate the object to be further away from the camera.

\hspace{-1.9em}
\begin{table}
\caption{Robot-to-object positioning error as a function of distance ($d_i$) from the camera to the robot, using various YCB objects.} 
\label{tab:expdist_ycb}
\centering
\begin{small}
\begin{tabular}{cccccccccc}
\toprule
 & \multicolumn{7}{c}{mean error (cm), $4$ object positions} \\
 \cmidrule(r){2-8} 
object & $d_1$ & $d_2$ & $d_3$ & $d_4$ & $d_5$ & $d_6$ & $d_7$ \\
\midrule
sugar box & 0.6 & 0.7 & 0.5 & 1.3 & 1.3 & 1.1 & 1.9 \\
cracker box & 1.6 & 2.4 & 2.0 & 2.6 & 2.4 & 1.7 & 3.0 \\
soup can & 1.0 & 0.5 & 0.9 & 1.3 & 0.4 & 1.2 & 1.5 \\
meat can & 1.3 & 1.1 & 1.0 & 1.8 & 1.4 & 1.6 & 2.3 \\
mustard & 1.5 & 3.8 & 3.9 & 3.3 & 6.4 & 5.8 & 3.9 \\
\bottomrule
\end{tabular}
\end{small}
\end{table}

In the previous experiments, we used only a favorable object orientation while varying camera distance.
In the next experiment, we maintained a constant distance ($d=112$~cm) and object position while varying the pose of each object at roughly $10^\circ$ increments in an upright position, along with a couple other orientations (\eg side, upside down), leading to $20$--$25$ measurements per object.
As before, we tabulated the lateral distance between the object centroid and end effector using the hanging weight.
Tab.~\ref{tab:exporient} presents the mean and standard deviation of the readings for YCB objects of three representative shapes (cuboid, cylinder, and bottle).
We maintained the same system configuration as the previous experiment (\viz last row in Tab.~\ref{tab:expdist}).
Other than the mustard bottle, which again due to lack of texture yielded worse performance compared to the other objects, accuracy (even including one standard deviation) was mostly within the grasping threshold.

\begin{table}
\centering
\caption{Robot-to-object positioning error over multiple object orientations, using three YCB objects.} 
\label{tab:exporient}
\begin{small}
\begin{tabular}{cccccccccc}
\toprule
object & mean $\pm$ std error (cm) \\
\midrule
sugar box & 1.5 $\pm$ 0.6 \\
soup can & 1.2 $\pm$ 0.7 \\
mustard & 1.9 $\pm$ 0.9 \\
\bottomrule
\end{tabular}
\end{small}
\end{table}

Finally we measured positioning accuracy on all $28$ proposed new objects from a fixed camera distance ($d=112$~cm).
We used the same networks as above, but did not use a refinement step, to better directly understand the accuracy of the combined neural networks.
Tab.~\ref{tab:exphope} presents the results of placing each object in four distinct locations in front of the robot, as before, with favorable orientation.
The table is organized by object shape (cuboids, cylinders, and bottles).
From the results, cuboids are the most reliable, if the spaghetti box is ignored; this object is uniquely challenging due its long and thin shape. 
Some of the cylinders are also quite difficult due to their small size, and the bottles present similar challenges as the YCB mustard bottle due to lack of texture.
Another issue is what we call pseudo-symmetry, where only a tiny detail (such as a lip on a box) differentiates two symmetric configurations, \eg the front and back of the cookie box.
From these results, we believe that these objects will be useful to the robotics and computer vision communities (given that some objects are well detected and stable), as well as challenging for future research (since many objects yield poor accuracy, with errors above $2$~cm).

\begin{table}
\centering
\caption{Robot-to-object positioning error using toy grocery objects.} 
\label{tab:exphope}
\begin{small}
\begin{tabular}{rcccccccccc}
\toprule
& object & dim. (cm) & mean $\pm$ std error (cm)\\
\midrule

\multirow{10}{1em}{\rotatebox{90}{CUBOIDS}} 
& choc.~pudding & \hspace{0.2em} 8.3 & 3.1 $\pm$ 0.6 \\
& butter & 10.3 & 1.3 $\pm$ 1.3 \\
& cream cheese & 10.4 & 2.5 $\pm$ 0.6 \\
& raisins & 12.3 & 3.8 $\pm$ 0.9 \\
& popcorn & 12.6 & 1.1 $\pm$ 0.4 \\
& granola bars & 16.5 & 1.8 $\pm$ 1.2 \\
& mac.~\& cheese & 16.6 & 1.5 $\pm$ 0.7 \\
& cookies & 16.7 & 1.3 $\pm$ 0.9 \\
& spaghetti & 25.0 & 6.8 $\pm$ 1.3 \\
& \emph{average} & -- & \emph{2.6 $\pm$ 0.9} \\

\midrule  %
\multirow{13}{1em}{\rotatebox{90}{CYLINDERS}} 
& yogurt & \hspace{0.2em} 6.8 & 2.0 $\pm$ 0.8 \\
& pineapple & \hspace{0.2em} 7.0 & 1.1 $\pm$ 0.3 \\
& tomato sauce & \hspace{0.2em} 7.0 & 1.1 $\pm$ 0.1 \\
& corn & \hspace{0.2em} 7.1 & 1.3 $\pm$ 0.7 \\
& green beans & \hspace{0.2em} 7.1 & 3.2 $\pm$ 2.1 \\
& mushrooms & \hspace{0.2em} 7.1 & 3.1 $\pm$ 0.9 \\
& peaches & \hspace{0.2em} 7.1 & 3.0 $\pm$ 0.6 \\
& peas \& carrots & \hspace{0.2em} 7.1 & 3.0 $\pm$ 0.8 \\
& tuna & \hspace{0.2em} 7.1 & 1.9 $\pm$ 1.2 \\
& alphabet soup & \hspace{0.2em} 8.4 & 1.8 $\pm$ 0.3 \\
& cherries & 10.3 & 5.7 $\pm$ 3.8 \\
& parmesan & 10.3 & 1.8 $\pm$ 0.2 \\
& \emph{average} & -- & \emph{2.4 $\pm$ 1.0} \\

\midrule  %
\multirow{8}{1em}{\rotatebox{90}{BOTTLES}} 
& salad dressing & 14.7 & 1.9 $\pm$ 0.7 \\
& mayo & 14.8 & 2.2 $\pm$ 0.4 \\
& bbq sauce & 14.8 & 3.5 $\pm$ 0.9 \\
& ketchup & 14.9 & 1.9 $\pm$ 0.2 \\
& mustard & 16.0 & 3.5 $\pm$ 1.0 \\
& milk & 19.0 & 1.8 $\pm$ 0.2 \\
& orange juice & 19.2 & 1.6 $\pm$ 0.3 \\
& \emph{average} & -- & \emph{2.3 $\pm$ 0.5} \\

\bottomrule
\end{tabular}
\end{small}
\end{table}

\subsection{Camera motion}

\begin{figure}
    \centering
    \begin{tabular}{c}
        \hspace{-0.5em}\includegraphics[width=\linewidth]{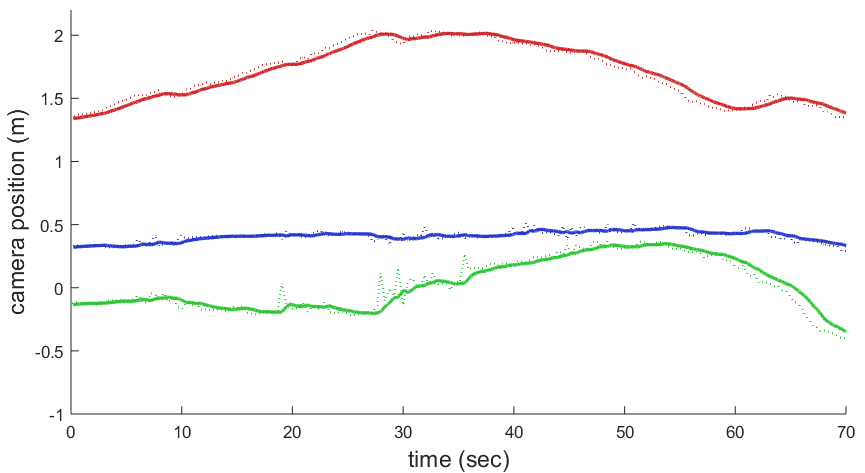} \\
        \hspace{-0.5em}\includegraphics[width=\linewidth]{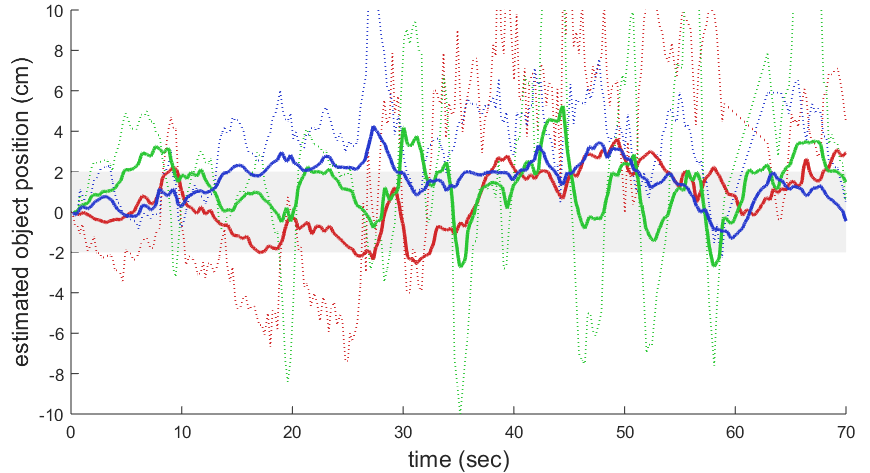} \\
        \hspace{-0.5em}\includegraphics[width=\linewidth]{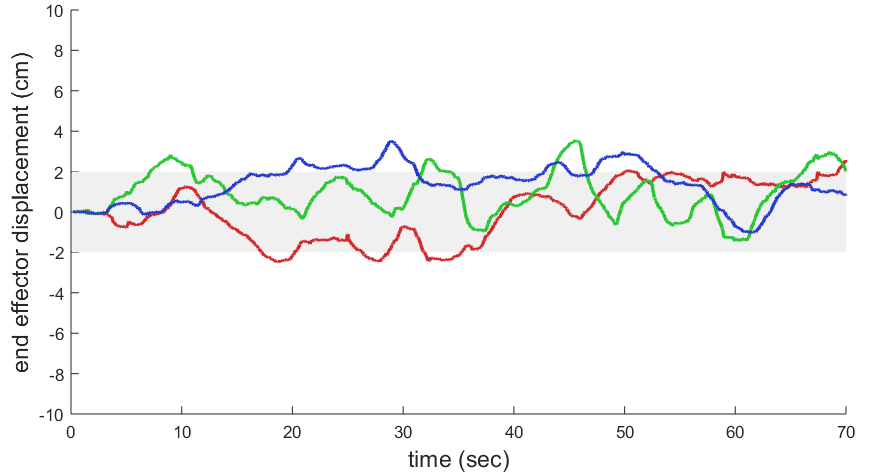}
    \end{tabular}
    \caption{As the handheld camera was moved around the scene (top), the object (middle) and end effector (bottom) poses remained fairly stable, as the robot hovered over the object.  Shown are the $xyz$ coordinates of the poses in the robot coordinate frame as red/green/blue, respectively.  The dashed and solid lines show raw and exponentially filtered poses, respectively.}
    \label{fig:campose}
\end{figure}

In this experiment we investigate the stability of the system when the camera is moved freely. 
The sugar box was placed on a table in front of the robot with the robot arm center above it.
The prior network configuration was used with pose refinement.
The camera was then manually moved around the room while ensuring that the robot and object remained in the field of view.
The first row of Fig.~\ref{fig:campose} shows the camera movement as perceived by our 
pose estimator, using the standard coordinate frames of the Baxter robot, \ie the $x$ axis (red curve) points toward the front of the robot, the $y$ axis (green curve) points left (which is approximately the camera's right), and the $z$ axis (blue curve) points up.
By this estimate, the camera was moved from a depth of $1.3$ to $2.0$~m and laterally from $-0.3$ to $0.3$~m while its height ranged from $0.3$ to $0.5$~m.

The second row of Fig.~\ref{fig:campose} shows the estimated object position relative to its starting position in the robot coordinate system; since the sugar box did not move, in a perfect system all colored curves should be horizontal lines at zero. 
With exponential filtering, the estimated position remains largely within or near the $2$~cm range of graspability for most of the sequence.
Because the sugar box was aligned to be parallel to the face plate of the robot, the $x$ direction (red curve) is most sensitive for grasping, the $y$ direction (green curve) is less sensitive, and the $z$ direction (blue curve) is the least important (since it is the approach direction).

The last row of the figure shows end effector displacement relative to its initial position as the closed-loop system attempted to hover over the object.
As before, in a perfect system all lines would be horizontal at zero.
The dynamics of the physical system dampens the errors slightly but also introduces slight errors of its own.

\begin{figure}
    \centering
    \begin{tabular}{c}
        \hspace{-0.5em}\includegraphics[width=0.8\linewidth]{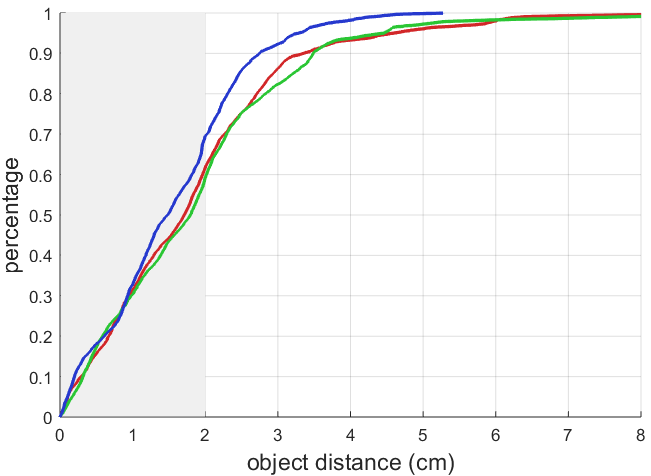} \\
    \end{tabular}
    \caption{The cumulative distribution function of object pose estimation for the experiment in Fig.~\ref{fig:campose}. 
    The lines cross $2$~cm (Baxter tolerance) at $0.62$ (red), $0.59$ (green), and $0.70$ (blue).  In other words, for approximately $60\%$ of the image frames, the error was within the grasp tolerance.}
    \label{fig:cdf}
\end{figure}

Fig.~\ref{fig:cdf} shows the cumulative distribution function for all three axes of object displacement from this experiment.
(The plot for end effector displacement is similar.)
Again, the most important axes are the red and green, in that order, since these are the directions in which the gripper has the least tolerance for error. 
Despite the unstructured movement of the camera and the use of non-full image resolution, the robot remains within the graspable region approximately $60\%$ of the time.
In other words, roughly $60\%$ of the time, the robot would have successfully grasped the object had it been commanded to move down and close the fingers.

\subsection{Grasping experiments}

\begin{figure*}
    \centering
    \begin{tabular}{ccccc}
        \hspace{-1em}
				\includegraphics[width=0.175\linewidth]{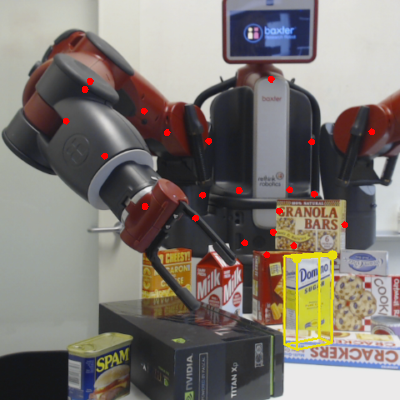} & 
        \includegraphics[width=0.175\linewidth]{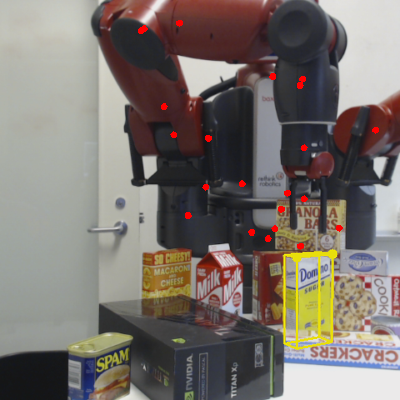} & 
        \includegraphics[width=0.175\linewidth]{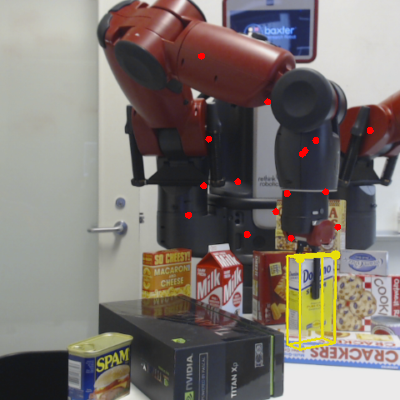} & 
        \includegraphics[width=0.175\linewidth]{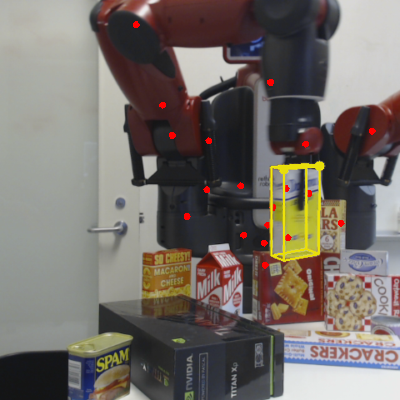} & 
        \includegraphics[width=0.175\linewidth]{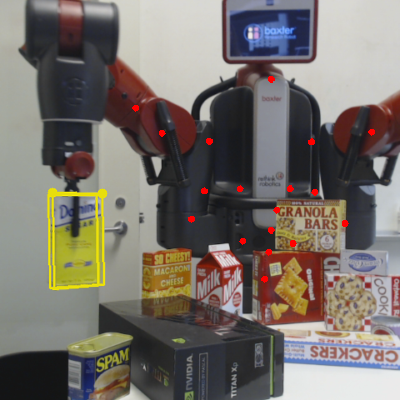}
    \end{tabular}
    \caption{Sequence of image frames from a successful grasp of the sugar box, using a camera approximately $1.6$~m from the robot.}
    \label{fig:grasping}
\end{figure*}

\begin{table}
\centering
\caption{Grasp success using various objects.} 
\label{tab:grasp}
\begin{small}
\begin{tabular}{rccccccccc}
\toprule
object & refinement & success \\
\midrule
sugar box & \checkmark  & 100\%  \\
soup can & \checkmark  &  80\% \\
mustard & \checkmark  & 90\%  \\
cracker box& \checkmark  & 100\%  \\
meat can &  \checkmark &  80\% \\
\midrule
milk & & 80\% \\ %
cookies &  & 90\% \\
popcorn &  & 90\% \\ 
tomato sauce &  & 100\% \\
parmesan & & 80\% \\
\midrule 
\emph{all} & & \emph{91\%}\\
\bottomrule
\end{tabular}
\end{small}
\end{table}

The ultimate test of the presented method is whether its accuracy is sufficient for robotic grasping.
As such, we designed a simple controller to move the robot gripper above the detected object, 
descend on it, and close the gripper; if the grasp was successful, the robot then picked up the object and dropped it in a basket. 
For each object we performed grasps at random locations in the workspace, including while stacked on other objects. 
Specifically, we performed $5$ grasps at $2$ different camera locations, yielding $10$ grasps per object. 
Both networks ran continuously during the manipulation; thus, the camera calibration was never fixed.
For this experiment we used the same network parameters as before. 

Table~\ref{tab:grasp} shows results using pose refinement for various YCB objects and without refinement for the proposed toy objects. 
Since Baxter's parallel jaw gripper has a rather small tolerance of $2$~cm when grasping an object (as mentioned earlier), we believe these results show the promise of such an approach in real robotics applications.
Fig.~\ref{fig:grasping} shows the robot successfully grasping an object using the proposed system; note that the object was continually detected, even while held by the gripper, which could be used in the future to detect slippage.

\section{DISCUSSION \& CONCLUSION}

In this work we have presented a system that computes object poses with respect to the robot indirectly, by combining the output of a neural network that estimates the object-to-camera pose with the output of another network that estimates the robot-to-camera pose.
The system uses a single external RGB camera as input, which can be moved freely during operation without significantly affecting accuracy, due to the online calibration provided by the latter network.
We have shown that our system is accurate enough to perform robotic grasps with $2$~cm tolerance.
We also introduce a new set of toy grocery objects that are easily accessible to other researchers due to their availability online.
To aid reproducibility of our results, we share the 3D scanned textured models and pretrained weights for pose estimation of these novel objects.

It is difficult to convey in a paper the tremendous freedom that online calibration provides.
As researchers, in the past we have always been very careful not to accidentally nudge a camera after calibration, lest the system break, and a tedious recalibration step be required.
With this system, however, as we were running experiments, we were able to freely grab the camera and move it around, without fear that the system would stop working.
To be fair, the exponential filtering is critical for safe performance; but with this filtering we can pick up and move the object or camera, walk around the robot, occlude the camera, and so forth, all while the robot is servoing on the object position.
In this respect, the performance we observe is remarkably stable.

Having said that, the problem is far from being solved.
Many aspects of the system can cause grasp failure:  incorrect size of the 3D scanned textured model, inaccurate pose estimation from either network due to occlusion or lighting conditions, pose refinement getting stuck in a local minimum, the object being aligned with the camera image plane so that only one face is visible (in the case of a cuboid, which increases error), and so forth.
Moreover, for object detection, false negatives occasionally prevent the object's pose from being estimated, which we noticed particularly as the lighting conditions of the room changed from morning to evening; rarely, false positives can cause the robot to wander to strange places in the scene (though this is largely mitigated by the exponential filtering).

For future work, there are many ways to improve the robustness of the proposed system, including using higher-resolution input images and output belief maps, incorporating tracking into the state estimation, fine-tuning the networks with real data, and improving the quality of the 3D scanned textured models used.
Moreover, achieving reliable grasping with a single external RGB camera is inherently difficult due to the lack of depth information.
As a result, some combination of depth, stereo, and/or camera-in-hand may be necessary to achieve mission-critical levels of robustness.

\section*{ACKNOWLEDGMENTS}

The authors are indebted to Timothy Lee, Thang To, Jia Cheng, Jeffrey Smith, Christopher Paxton, and Clemens Eppner for their help.

\bibliographystyle{IEEEtran}
\bibliography{refs}

\begin{thebibliography}{10}
\providecommand{\url}[1]{#1}
\csname url@rmstyle\endcsname
\providecommand{\newblock}{\relax}
\providecommand{\bibinfo}[2]{#2}
\providecommand\BIBentrySTDinterwordspacing{\spaceskip=0pt\relax}
\providecommand\BIBentryALTinterwordstretchfactor{4}
\providecommand\BIBentryALTinterwordspacing{\spaceskip=\fontdimen2\font plus
\BIBentryALTinterwordstretchfactor\fontdimen3\font minus
  \fontdimen4\font\relax}
\providecommand\BIBforeignlanguage[2]{{%
\expandafter\ifx\csname l@#1\endcsname\relax
\typeout{** WARNING: IEEEtran.bst: No hyphenation pattern has been}%
\typeout{** loaded for the language `#1'. Using the pattern for}%
\typeout{** the default language instead.}%
\else
\language=\csname l@#1\endcsname
\fi
#2}}

\bibitem{xiang2018rss:posecnn}
Y.~Xiang, T.~Schmidt, V.~Narayanan, and D.~Fox, ``Pose{CNN}: {A} convolutional
  neural network for {6D} object pose estimation in cluttered scenes,'' in
  \emph{RSS}, 2018.

\bibitem{tremblay2018corl:dope}
J.~Tremblay, T.~To, B.~Sundaralingam, Y.~Xiang, D.~Fox, and S.~Birchfield,
  ``Deep object pose estimation for semantic robotic grasping of household
  objects,'' in \emph{CoRL}, 2018.

\bibitem{peng2019pvnet}
S.~Peng, Y.~Liu, Q.~Huang, X.~Zhou, and H.~Bao, ``{PVNet}: {P}ixel-wise voting
  network for {6DoF} pose estimation,'' in \emph{CVPR}, 2019.

\bibitem{zakharov2019dpod}
S.~Zakharov, I.~Shugurov, and S.~Ilic, ``{DPOD}: {D}ense {6D} pose object
  detector in {RGB} images,'' in \emph{ICCV}, 2019.

\bibitem{hu2019segmentation}
Y.~Hu, J.~Hugonot, P.~Fua, and M.~Salzmann, ``Segmentation-driven {6D} object
  pose estimation,'' in \emph{CVPR}, 2019.

\bibitem{wang2019cvpr:densefus}
C.~Wang, D.~Xu, Y.~Zhu, R.~Martin-Martin, C.~Lu, L.~Fei-Fei, and S.~Savarese,
  ``{DenseFusion}: {6D} object pose estimation by iterative dense fusion,'' in
  \emph{CVPR}, 2019.

\bibitem{fiala2005cvpr:artag}
M.~Fiala, ``{ARTag}, a fiducial marker system using digital techniques,'' in
  \emph{CVPR}, 2005.

\bibitem{olson2011icra:apriltags}
E.~Olson, ``{AprilTag}: A robust and flexible visual fiducial system,'' in
  \emph{ICRA}, 2011.

\bibitem{lee2020icra:dream}
T.~E. Lee, J.~Tremblay, T.~To, J.~Cheng, T.~Mosier, O.~Kroemer, D.~Fox, and
  S.~Birchfield, ``Camera-to-robot pose estimation from a single image,'' in
  \emph{ICRA}, 2020.

\bibitem{tobin2017iros:dr}
J.~Tobin, R.~Fong, A.~Ray, J.~Schneider, W.~Zaremba, and P.~Abbeel, ``Domain
  randomization for transferring deep neural networks from simulation to the
  real world,'' in \emph{IROS}, 2017.

\bibitem{wang2020icra:6pack}
C.~Wang, R.~Mart{\'\i}n-Mart{\'\i}n, D.~Xu, J.~Lv, C.~Lu, L.~Fei-Fei,
  S.~Savarese, and Y.~Zhu, ``{6-PACK}: {C}ategory-level {6D} pose tracker with
  anchor-based keypoints,'' in \emph{ICRA}, 2020.

\bibitem{stevsic2020ral:ltass}
S.~Stev\v{s}i\'{c}, S.~Christen, and O.~Hilliges, ``Learning to assemble:
  {E}stimating {6D} poses for robotic object-object manipulation,'' \emph{IEEE
  Robotics and Automation Letters}, vol.~5, no.~2, pp. 1159--1166, 2020.

\bibitem{florencemanuelli2018dense}
P.~Florence, L.~Manuelli, and R.~Tedrake, ``Dense object nets: {L}earning dense
  visual object descriptors by and for robotic manipulation,'' in
  \emph{Conference on Robot Learning (CoRL)}, 2018.

\bibitem{manuelli2019kpam}
L.~Manuelli, W.~Gao, P.~Florence, and R.~Tedrake, ``{kPAM}: {KeyPoint}
  affordances for category-level robotic manipulation,'' in \emph{ISRR}, 2019.

\bibitem{tremblay2018icra:cube}
J.~Tremblay, T.~To, A.~Molchanov, S.~Tyree, J.~Kautz, and S.~Birchfield,
  ``Synthetically trained neural networks for learning human-readable plans
  from real-world demonstrations,'' in \emph{ICRA}, 2018.

\bibitem{mahler2017rss:dexnet2}
J.~Mahler, J.~Liang, S.~Niyaz, M.~Laskey, R.~Doan, X.~Liu, J.~A. Ojea, and
  K.~Goldberg, ``{Dex-Net} 2.0: {D}eep learning to plan robust grasps with
  synthetic point clouds and analytic grasp metrics,'' in \emph{RSS}, 2017.

\bibitem{Mousavian_2019_ICCV}
A.~Mousavian, C.~Eppner, and D.~Fox, ``{6-DOF GraspNet}: Variational grasp
  generation for object manipulation,'' in \emph{ICCV}, 2019.

\bibitem{murali2020icra:cluttergrasping}
A.~Murali, A.~Mousavian, C.~Eppner, C.~Paxton, and D.~Fox, ``{6-DOF} grasping
  for target-driven object manipulation in clutter,'' in \emph{ICRA}, 2020.

\bibitem{pauwels2016icra:onlinecalib}
K.~Pauwels and D.~Kragic, ``Integrated on-line robot-camera calibration and
  object pose estimation,'' in \emph{ICRA}, 2016.

\bibitem{james2017corl:transferring}
S.~James, A.~J. Davison, and E.~Johns, ``Transferring end-to-end visuomotor
  control from simulation to real world for a multi-stage task,'' in
  \emph{CoRL}, 2017.

\bibitem{rusu2017corl:prog}
A.~A. Rusu, M.~Vecerik, T.~Rothörl, N.~Heess, R.~Pascanu, and R.~Hadsell,
  ``Sim-to-real robot learning from pixels with progressive nets,'' in
  \emph{CoRL}, 2017.

\bibitem{lee2020icra:guapo}
M.~Lee, C.~Florensa, J.~Tremblay, N.~Ratliff, A.~Garg, F.~Ramos, and D.~Fox,
  ``Guided uncertainty-aware policy optimization: Combining model-free and
  model-based strategies for sample-efficient learning,'' in \emph{ICRA}, 2020.

\bibitem{smith2019avid}
L.~Smith, N.~Dhawan, M.~Zhang, P.~Abbeel, and S.~Levine, ``{AVID}: {L}earning
  multi-stage tasks via pixel-level translation of human videos,'' in
  \emph{NeurIPS Deep Reinforcement Learning Workshop}, 2019.

\bibitem{iqbal2020icra:directional}
S.~Iqbal, J.~Tremblay, T.~To, J.~Cheng, E.~Leitch, A.~Campbell, K.~Leung,
  D.~McKay, and S.~Birchfield, ``Toward sim-to-real directional semantic
  grasping,'' in \emph{ICRA}, 2020.

\bibitem{lepetit2009ijcv:epnp}
V.~Lepetit, F.~Moreno-Noguer, and P.~Fua, ``{EP\emph{n}P}: {A}n accurate
  {O}(\emph{n}) solution to the {P\emph{n}P} problem,'' \emph{IJCV}, vol.~81,
  no.~2, 2009.

\bibitem{simonyan2015iclr:vgg}
K.~Simonyan and A.~Zisserman, ``Very deep convolutional networks for
  large-scale image recognition,'' in \emph{ICLR}, 2015.

\bibitem{to2018ndds}
T.~To, J.~Tremblay, D.~McKay, Y.~Yamaguchi, K.~Leung, A.~Balanon, J.~Cheng, and
  S.~Birchfield, ``{NDDS}: {NVIDIA} deep learning dataset synthesizer,'' 2018,
  \url{https://github.com/NVIDIA/Dataset_Synthesizer}.

\bibitem{tremblay2018arx:fat}
J.~Tremblay, T.~To, and S.~Birchfield, ``Falling things: {A} synthetic dataset
  for {3D} object detection and pose estimation,'' in \emph{CVPR Workshop on
  Real World Challenges in Robotic Vision}, June 2018.

\bibitem{kingma2015iclr:adam}
D.~P. Kingma and J.~Ba, ``Adam: {A} method for stochastic optimization,'' in
  \emph{ICLR}, 2015.

\bibitem{jang2017corl:semgrasp}
E.~Jang, S.~Vijayanarasimhan, P.~Pastor, J.~Ibarz, and S.~Levine, ``End-to-end
  learning of semantic grasping,'' in \emph{CoRL}, 2017.

\bibitem{xu2018icra:ntp}
D.~Xu, S.~Nair, Y.~Zhu, J.~Gao, A.~Garg, L.~Fei-Fei, and S.~Savarese, ``Neural
  task programming: {L}earning to generalize across hierarchical tasks,'' in
  \emph{ICRA}, 2018.

\bibitem{levine2016jmlr:ete}
S.~Levine, C.~Finn, T.~Darrell, and P.~Abbeel, ``End-to-end training of deep
  visuomotor policies,'' \emph{Journal of Machine Learning Research (JMLR)},
  2016.

\bibitem{zhu2018rss:rlildvs}
Y.~Zhu \emph{et~al.}, ``Reinforcement and imitation learning for diverse
  visuomotor skills,'' in \emph{RSS}, 2018.

\bibitem{pauwels2016csvt:posetrack}
K.~Pauwels, L.~Rubio, and E.~Ros, ``Real-time pose detection and tracking of
  hundreds of objects,'' \emph{IEEE Transactions on Circuits and Systems for
  Video Technology}, vol.~26, no.~12, Dec. 2016.

\bibitem{pauwels2015iros:simtrack}
K.~Pauwels and D.~Kragic, ``{SimTrack}: {A} simulation-based framework for
  scalable real-time object pose detection and tracking,'' in \emph{IROS},
  2015.

\bibitem{calli2015ram:ycb}
B.~Calli, A.~Walsman, A.~Singh, S.~Srinivasa, P.~Abbeel, and A.~M. Dollar,
  ``Benchmarking in manipulation research: {U}sing the {Yale-CMU-Berkeley}
  object and model set,'' \emph{IEEE Robotics and Automation Magazine},
  vol.~22, no.~3, Sept. 2015.

\bibitem{ratliff2018arx:rmp}
N.~D. Ratliff, J.~Issac, D.~Kappler, S.~Birchfield, and D.~Fox, ``Riemannian
  motion policies,'' \emph{arXiv:1801.02854}, 2018.

\bibitem{cheng2018wafr:rmpflow}
C.-A. Cheng, M.~Mukadam, J.~Issac, S.~Birchfield, D.~Fox, B.~Boots, and
  N.~Ratliff, ``{RMPflow}: {A} computational graph for automatic motion policy
  generation,'' in \emph{WAFR}, 2018.

\end{thebibliography}

\addtolength{\textheight}{-12cm}   %

\end{document}